\documentclass[letterpaper, 10 pt, conference]{ieeeconf}  %

\IEEEoverridecommandlockouts                              %

\usepackage{graphics} %
\usepackage{graphicx}
\usepackage{subcaption}

\usepackage{bm}

\usepackage[noadjust]{cite}

\usepackage{hyperref}
\usepackage{amsmath} %
\usepackage{amssymb}  %
\usepackage[ruled,linesnumbered,vlined]{algorithm2e}
\usepackage{booktabs}
\usepackage{multirow}
\usepackage{cleveref}
\usepackage{color, soul}
\usepackage{array}
\usepackage{makecell}
\usepackage{bigstrut}
\usepackage{authblk}
\newcolumntype{C}[1]{>{\centering\arraybackslash}p{#1}}

\title{\LARGE \bf
  3D Surfel Map-Aided Visual Relocalization with Learned Descriptors
}

\def\thankstextone{
$^{1}$ H. Ye, H. Huang and M. Liu are with \href{https://ram-lab.com/}{RAM-LAB}, the Hong Kong University of Science and Technology, Kowloon, Hong Kong.
}

\def\thankstexttwo{
$^{2}$ M. Hutter and T. Sandy are with Robotic Systems Lab, ETH Zurich, Switzerland.
}

\author{Haoyang~Ye$^{1}$, Huaiyang~Huang$^{1}$, Marco Hutter$^{2}$, Timothy Sandy$^{2}$, and~Ming~Liu$^{1}$%
\thanks{This work was supported by the National Natural Science Foundation of China, under grant No. U1713211, Collaborative Research Fund by Research Grants Council Hong Kong, under Project No. C4063-18G, and HKUST-SJTU Joint Research Collaboration Fund, under project SJTU20EG03, awarded to Prof. Ming Liu.}
\thanks{\thankstextone}%
\thanks{\thankstexttwo}%
}

\begin{document}

\maketitle
\thispagestyle{empty}
\pagestyle{empty}

\begin{abstract}
  In this paper, we introduce a method for visual relocalization using the geometric information from a 3D surfel map.
  A visual database is first built by global indices from the 3D surfel map rendering, which provides associations between image points and 3D surfels.
  Surfel reprojection constraints are utilized to optimize the keyframe poses and map points in the visual database.
  A hierarchical camera relocalization algorithm then utilizes the visual database to estimate 6-DoF camera poses.
  Learned descriptors are further used to improve the performance in challenging cases.
  We present evaluation under real-world conditions and simulation to show the effectiveness and efficiency of our method, and make the final camera poses consistently well aligned with the 3D environment.
\end{abstract}

\setlength{\textfloatsep}{0.5pt}

\section{Introduction}
Relocalization is one of the essential requirements for autonomous robots.
When a navigation system starts or the estimation of the robot pose is lost, the relocalization module is required to provide an initial six-degrees-of-freedom (6-DoF) pose for the system.
In many visual localization systems \cite{caselitz2016monocular,huang2019metric,ye2020monocular}, an initial pose is assumed to be known in achieving the subsequent camera tracking.
Thus, how to obtain the initial pose in a 3D environment becomes an important problem.

A GNSS/IMU system is typically used for outdoor applications to provide the initial poses.
However, in GNSS-denied environments, relocalization with only on-board sensors is necessary.
Cameras are one of the most popular sensors for autonomous robots, thanks to their low cost and light weight, and many visual simultaneous localization and mapping (SLAM) systems \cite{mur2015orb,campos2020orb} take sequential camera data as input to locate the camera as well as to build the map of the surroundings.
However, drift is inevitable in visual SLAM systems, causing gaps between the real 3D model and the mapped one and degrading the metric relocalization performance.
Structure from motion (SfM) \cite{schoenberger2016sfm} provides more accurate visual structure, and is utilized in many visual (re-)localization-related works \cite{irschara2009structure,li2012worldwide,sarlin2018leveraging,sarlin2019coarse}.
However, SfM becomes inefficient and time-consuming when the number of input images increases.
It is also not flexible to update the unmapped areas of the SfM model with new images.

To efficiently model the 3D structure, 3D lidars are a good choice since they can provide accurate distance measurements and map 3D environments in real time \cite{zhang2014loam,shan2018lego,ye2019tightly}.
Maps built by 3D lidars are usually of higher quality compared to those from visual SLAM systems.
Although 3D lidars are expensive and bulky, which may not suit low-cost platforms, leveraging the pre-built 3D lidar map to improve the camera relocalization capability is still promising.
One of the most straightforward ways of utilizing the 3D information for relocalization is to directly associate the 3D map lidar points and 2D image features from images by given camera poses.
However, in this way, no inter-frame information, e.g., the covisibility of the frames or the trackability of the points in the database, is considered.
The covisibility of the frames helps to find more co-observed 3D structures beyond the most similar frame from the database \cite{sarlin2019coarse}, while the trackability of the points helps to keep trackable 3D points among frames and filter unreliable 3D points.
This information can help to improve the recall and accuracy of visual relocalization.

In this paper, we propose an efficient method to take the geometric information from the 3D surfel map to build a visual database for 6-DoF camera relocalization.
It takes a 3D surfel map, image sequences, and the corresponding camera poses in the 3D map as inputs.
The image points are associated with the 3D surfels and form a visual database, which can be further optimized by surfel reprojection constraints.
Within the same surfel map, the visual database can be further updated by new image sequences and camera poses.
The main contributions of our paper are:
\begin{itemize}
    \item An efficient way to build and update a visual database using geometric information from a 3D surfel map, which makes the visual database consistent with the 3D geometry.
    \item Proposed surfel reprojection constraints to further optimize the visual database.
    By considering surfel information from the 3D model, keyframes and map points are globally optimized, which improves the final relocalization recall and pose accuracy.
    \item A hierarchical relocalization strategy utilizing the geometric information from the 3D map, which retrieves the global closest frames and utilizes the covisible and map point information from the visual database to estimate 6-DoF camera poses.
    \item A validation and ablation study in both simulation and real-world experiments to show the effectiveness of the proposed method.
\end{itemize}

\section{Related Work}

In this section, we review the work related to visual (re-)localization.
That is, given one or few sequential images, camera poses in the visual database are estimated.
Based on the different types of visual databases, visual (re-)localization can be divided into three categories, 2D image-based, 3D structure-based and hierarchical approaches \cite{sattler2018benchmarking,sarlin2019coarse}.

2D image-based approaches provide approximate camera poses which correspond to the most similar images in the database.
Vector of locally aggregated descriptors (VLAD) \cite{jegou2010aggregating} and bag of words (BoW) \cite{sivic2003video} are two commonly used compact representations in 2D image-based approaches.
These approaches are most related to solving place recognition or image retrieval problems \cite{arandjelovic2016netvlad,torii201524,sattler2017large,cummins2008fab,galvez2012bags}.
They offer efficient database construction \cite{sattler2017large}, but cannot accurately provide 6-DoF camera poses.

3D structure-based approaches assume a 3D model of the environment is available.
The model can be a structure from motion (SfM) model \cite{li2012worldwide,svarm2016city,svarm2014accurate,zeisl2015camera,sattler2016efficient}, in which descriptor matching is applied to find associations between 2D image features and 3D map points.
Recently proposed scene coordinate regression networks \cite{brachmann2017dsac,zhou2020kfnet} use convolutional neural networks (CNNs) instead of SfM models to present the environments, and provide per-pixel 2D-3D matches.
After finding the 2D-3D matches, the camera poses can be estimated by Perspective-n-Point (PnP) solvers \cite{lepetit2009epnp,kneip2011novel} inside a RANSAC loop \cite{fischler1981random}.
3D structure-based approaches provide more accurate 6-DoF camera poses than 2D image-based approaches \cite{sattler2017large,sattler2018benchmarking,sattler2019understanding}, but updating the SfM model or networks with additional image inputs is not efficient in these approaches since re-building of the SfM model or re-training of the network is required.

Hierarchical approaches combine the advantages of 2D image-based and 3D structure-based approaches.
In \cite{irschara2009structure,middelberg2014scalable,sarlin2018leveraging} and \cite{sarlin2019coarse}, the visual relocalization problem is divided into two steps.
In the first step, global retrieval based on 2D image-based approaches is applied.
It provides a smaller search region for 2D-3D matching for the next step, where camera poses are estimated similar to the 3D structure-based approaches.

These hierarchical approaches, however, still require a visual database built from SfM \cite{schoenberger2016sfm,schoenberger2016mvs}, which is not flexible to update and typically slow to build.
To build the visual database for relocalization efficiently, visual SLAM systems \cite{mur2015orb,campos2020orb} update the visual database continuously since they process the images progressively instead of using all images at once.
However, camera pose drift in the SLAM system could cause the visual database to be inconsistent with the true 3D structure.

Based on this observation, our method leverages a 3D surfel map built from 3D lidars, which can be built more efficiently and densely than those from visual-based methods.
By utilizing the 3D surfel map, our method makes the visual database consistent with 3D environments.
Meanwhile, the visual database can be continuously updated with additional image inputs without re-building the whole database.

\section{Method}

Our method contains three major steps.
The first step (Sec. \ref{sec:database}) is to utilize image inputs, camera trajectory and a surfel map to build a visual database that is consistent with the 3D environment.
Then, surfel reprojection constraints are utilized to optimize the keyframe poses and map point locations (Sec. \ref{sec:surfel_reproj_opt}).
Finally, with the optimized visual database, we use hierarchical relocalization to estimate 6-DoF camera poses (Sec. \ref{sec:hierarchical_relocal}).

\begin{figure}[!ht]
    \centering
    \begin{subfigure}[c]{0.45\textwidth}
        \centering
        \includegraphics[width=\textwidth]{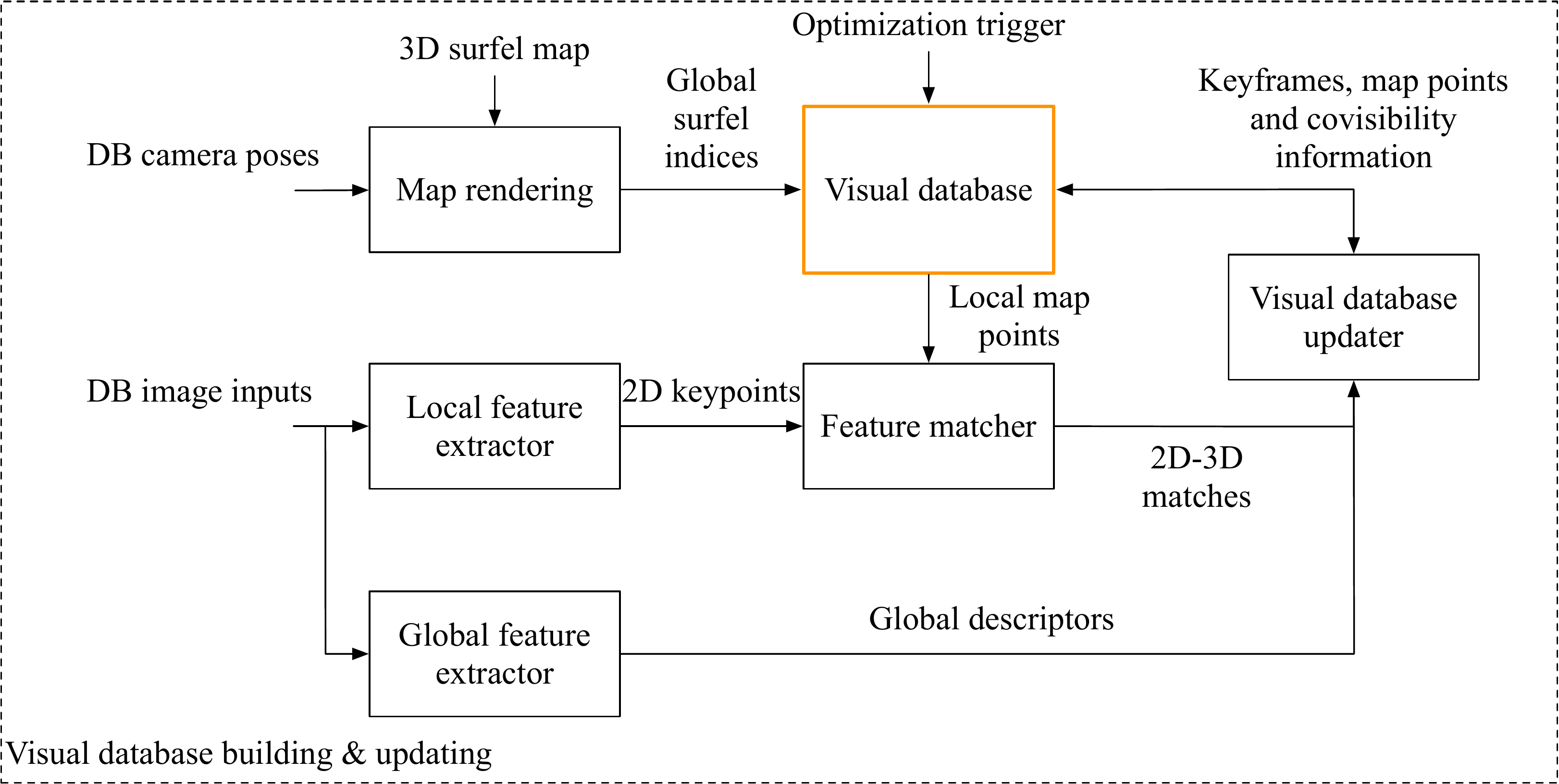}
        \caption{Pipeline of visual database building \& updating (Sec. \ref{sec:database} and \ref{sec:surfel_reproj_opt}).}
        \label{fig:database_diagram}
    \end{subfigure}
    \begin{subfigure}[c]{0.45\textwidth}
        \centering
        \includegraphics[width=\textwidth]{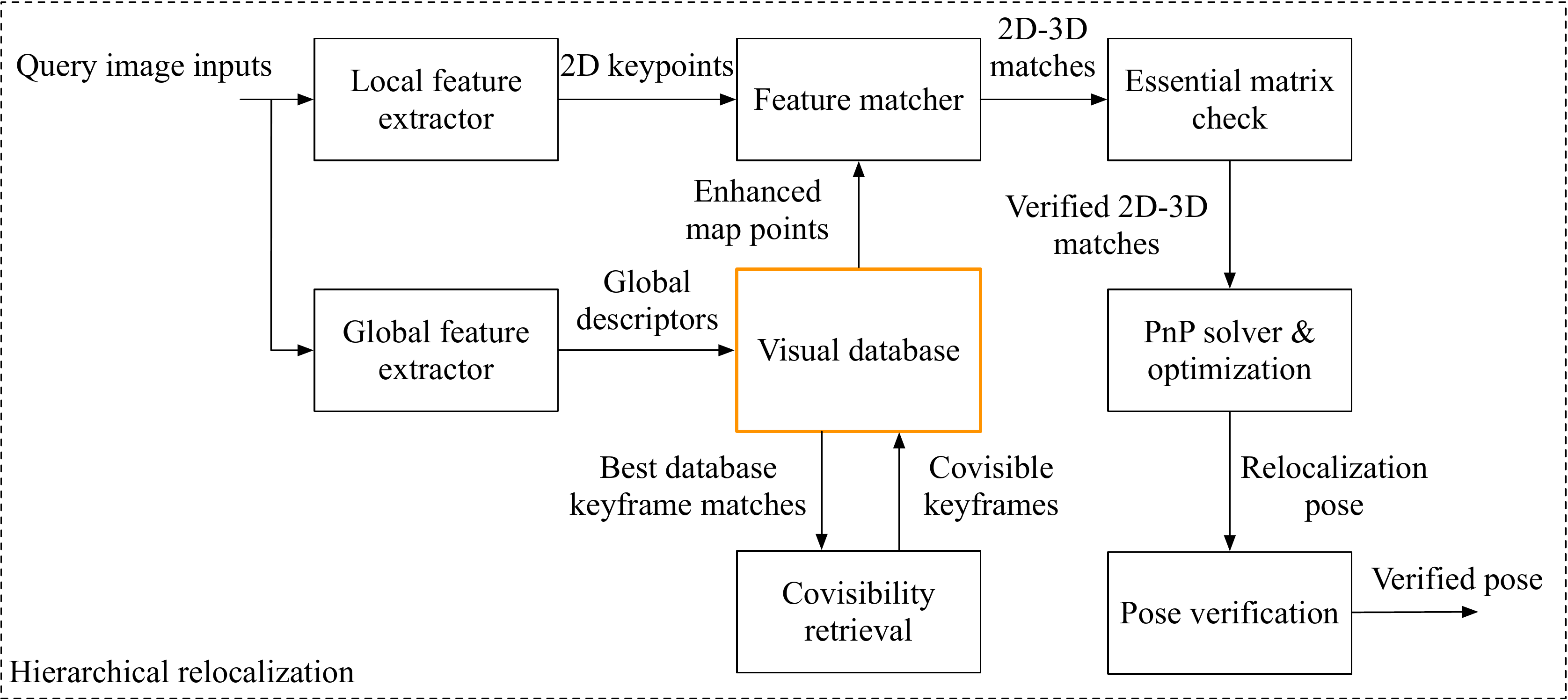}
        \caption{Pipeline of visual relocalization (Sec. \ref{sec:hierarchical_relocal}).}
        \label{fig:relocalization_diagram}
    \end{subfigure}
    \caption{
        Pipelines for building visual database and visual relocalization.
        (a) Sequential image inputs and their corresponding camera poses are utilized to build the visual database leveraging the geometric information from the 3D surfel map.
        After processing the image inputs, surfel reprojection optimization is triggered to improve the quality of the visual database.
        (b) When query images arrive, global and local descriptors are extracted for keyframe retrieval and 6-DoF camera pose estimation, respectively.
    }
    \vspace{-1.75em}
\end{figure}

\subsection{Visual Database Building with a 3D Surfel Map} \label{sec:database}
As shown in Fig. \ref{fig:database_diagram}, in the visual database building step, we take image inputs, the corresponding camera poses and a 3D surfel map as inputs.
Usually, the camera poses in the 3D environment are not directly available.
So there are two practical ways to obtain the camera poses.
1) If an initial guess of the camera pose is available, we can use camera localization algorithms \cite{caselitz2016monocular,huang2019metric,ye2020monocular} to calculate the trajectory of the camera in the 3D map.
2) If the 3D map is built by a 3D lidar in motion and rigidly-fixed synchronized camera data is captured at the same time, the camera poses can be estimated by transforming the lidar poses from the lidar mapping algorithms \cite{zhang2014loam, ye2019tightly} with the extrinsic parameters between the lidar and camera.

When the \textit{visual database building \& updating} step starts, the map rendering module loads the surfel map \cite{whelan2015elasticfusion} once.
We use the same map rendering module as in our previous work \cite{ye2020monocular}.
Given a camera pose \(\mathbf{T}^W_{k}\) obtained in one of the above-mentioned two ways, the map rendering module renders a \textit{global surfel indices map} in the image plane at timestamp \(k\).
Each available pixel in the global surfel indices map is associated with one index of the global surfels in the 3D surfel map.
After obtaining the image feature points from the \textit{local feature extractor} module, we can associate 2D feature points \(\{ \mathbf{p} \}_k\) in image \(I_k\) and surfels \(\{ S \}_k\) in the global surfel indices map, where \(\{\cdot\}_k\) denotes a set of points or surfels at timestamp \(k\).
3D map points \(\{\mathbf{x}\}_k\) associated with \(\{ \mathbf{p} \}_k\) are created.
The descriptors of \(\{\mathbf{x}\}_k\) are set to the corresponding descriptors of \(\{ \mathbf{p} \}_k\), and the rough positions of \(\{\mathbf{x}\}_k\) are set to the center positions of \(\{S\}_k\).
Besides \(\{ S \}_k\), we also store neighboring surfel set \(\{ S \}^{n}_k\) within the keypoint size of \(\{ \mathbf{p} \}_k\).
\(\{ S \}^{n}_k\) is used in Sec. \ref{sec:hierarchical_relocal} to improve the relocalization recall.

With the next image \(I_{k+1}\) and camera pose \(\mathbf{T}^W_{k+1}\), \(\{S\}_{k+1}\) can be obtained in the same way.
All keyframes, except \(\mathcal{F}_{k+1}\), the keyframe at \(k+1\), observing surfel \(S \in \{S\}_{k+1}\) are collected as covisible frames \(\{\mathcal{F}\}^{co}_{k+1}\).
The covisible map points at \(k+1\) are defined as \( \{\mathbf{x} \}^{co}_{k+1} \), where \(\mathbf{x}\) can be observed by \(\mathcal{F}_\alpha\), \( \mathcal{F}_\alpha \in \{\mathcal{F}\}^{co}_{k+1} \).
We then project \( \{\mathbf{x}\}^{co}_{k+1} \) to \(I_{k+1}\) and match them with \(\{\mathbf{p}\}_{k+1}\).
Similar to \cite{mur2015orb}, the matching is performed in the local grid of \(I_{k+1}\) and Lowe’s ratio test \cite{lowe2004distinctive} is applied.
For each 2D-3D pair, we update the map point's descriptor by finding the best distinctive descriptor among the feature descriptors in \(\{\mathcal{F}\}^{co}_{k+1} \cup \mathcal{F}_{k+1}\).
For the non-matched 2D image points, we create new 3D map points \(\{\mathbf{x}\}_{k+1}\) by the corresponding surfels \(\{S\}_{k+1}\).

Since the camera may revisit the same environment during data collection, we want to bound the size of the keyframes and map points in one scene.
In our \textit{visual database updater}, if a frame contains 90\% of map points which are visible in at least three other keyframes, it is considered as a duplicate frame and culled from the visual database, and if new map points are not visible in the next two keyframes, we will remove them from the visual database.
This helps to keep map points likely to be re-observed among keyframes, preserving the best covisibility property.

As we need to retrieve database keyframes sharing the visual content to the query frame, we choose to embed the database keyframes to global descriptors, which are compact to store and efficient to compare.

\subsection{Visual Database Optimization Using Surfel Constraints} \label{sec:surfel_reproj_opt}
In the previous step, the map point positions are assigned to the surfel centers, which may not be accurate enough.
The camera trajectory, transformed from lidar poses with varying extrinsic parameters or time offsets, can also introduce uncertainty to the keyframe poses.
Since the keyframes and map points are associated through surfels, we can leverage this information to optimize the keyframe poses and map point positions.

\begin{figure}[]
    \centering
    \includegraphics[width=0.32\textwidth]{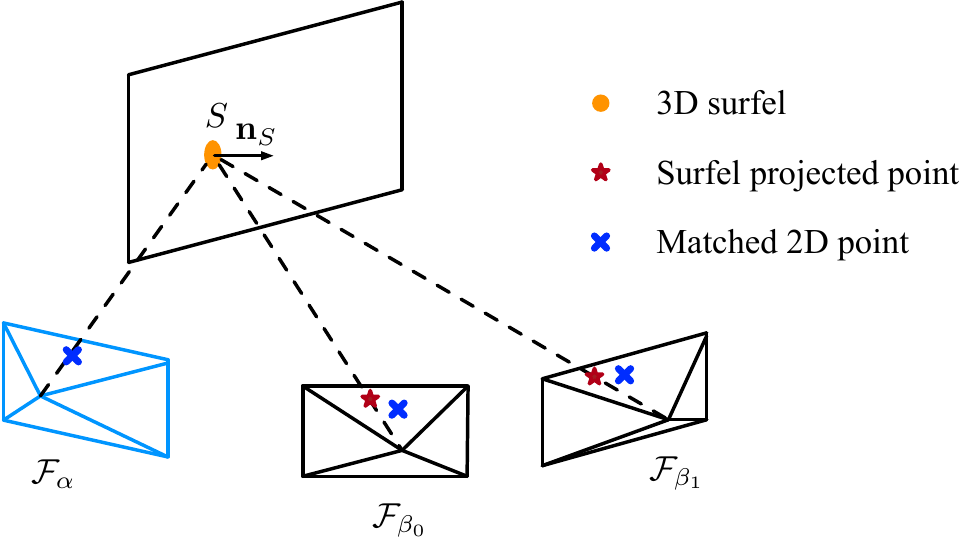}
    \caption{Reprojection induced by an associated surfel.
    \(\mathcal{F}_\alpha\) is the first keyframe observes the 3D surfel \(S\).
    Matched 2D points among the keyframes are denoted as blue crosses.
    The reprojected points in \(\mathcal{F}_{\beta_0}\) and  \(\mathcal{F}_{\beta_1}\) induced by the surfel are shown as red stars.
    }
    \label{fig:surfel_reprojection}
\end{figure}

Since each map point \(\mathbf{x}_W\) in the surfel map's frame \(\mathcal{F}_W\) is associated to a surfel \(S\), we can assume the map point is on a global surface with the same normal as the surfel \(\mathbf{n}_{S}\) and the center of surfel \(\mathbf{x}_{S}\) on the surface.
The plane coefficient can be written as \(\bm{\omega}_S = [\mathbf{n}_{S}\ d_{S} ]\), where \(d_{S} = -\mathbf{n}_{S}^\top \mathbf{x}_{S}\).
\(\mathbf{x}_W\) is observed by \(\mathcal{F}_k\) and \(\mathcal{F}_l\) in their image planes as \(\mathbf{p}_k\) and \(\mathbf{p}_l\), respectively.
The projection and unprojection functions transform \(\mathbf{x}_k\) in \(\mathcal{F}_k\) as \(\mathbf{p}_k = \Pi(\mathbf{x}_k)\) and \(\mathbf{x}_k = \Pi^{-1}(\mathbf{p}_k, \rho_k)\), respectively, where \(\rho_k\) is the inverse depth of \(\mathbf{x}_k\) in \(\mathcal{F}_k\).
Since \(\mathbf{x}_W\) is on the plane with coefficient \(\bm{\omega}_S\), it can be written as
\begin{equation} \label{eqn:world_point}
  \begin{gathered}
    \mathbf{x}_W^{\rho_k} = \mathbf{R}^W_k \cdot \Pi^{-1} \left(\mathbf{p}_k, \rho_k\right) + \mathbf{t}^W_k \\
    \mathbf{n}_{S}^\top \mathbf{x}_W^{\rho_k} + d_{S} = 0 
  \end{gathered},
\end{equation}
where \(\mathbf{T}^W_{(\cdot)}\) with \(\mathbf{R}^W_{(\cdot)}\) and \(\mathbf{t}^W_{(\cdot)}\) as its rotational and translational parts, respectively, transforms points in \(\mathcal{F}_{(\cdot)}\) to \(\mathcal{F}_W\).
The inverse depth \(\rho_k\) can be presented as
\begin{equation}
  \begin{aligned}
    \rho_k = - \frac{\mathbf{n}_S^\top \mathbf{R}^W_k \cdot \Pi_P^{-1}(\mathbf{p}_k)}{\mathbf{n}_S^\top \mathbf{t}^W_k + d_S}
  \end{aligned},
\end{equation}
where \( \Pi_P^{-1}(\cdot) \) lifts the image points to the unit image plane.
Then, the surfel reprojection error induced by the surfel \(S\) between two keyframes at timestamp \(k\) and \(l\) can be formulated as
\begin{equation} \label{eqn:surfel_reproj_res}
  \begin{aligned}
    \mathbf{e}_{sr}(S,k,l) = \Pi\left(
      (\mathbf{R}^W_l)^{-1} \cdot (\mathbf{x}_W^{\rho_k} - \mathbf{t}^W_l)
      \right) - \mathbf{p}_l
  \end{aligned}.
\end{equation}
Consider all surfels associated with keyframes.
The cost function among all surfels can be written as
\begin{equation} \label{eqn:surfel_reproj_cost} 
  \begin{aligned}
    c_{sr} = \sum_j \sum_{\alpha,\beta \in \text{obs}(S_j), \alpha \ne \beta} \left\lVert {\mathbf{e}_{sr} (S_j,\alpha,\beta)} \right\rVert^2
  \end{aligned},
\end{equation}
where \(S_j\) can be observed in \(\mathcal{F}_\alpha\) and \(\mathcal{F}_\beta\).
To reduce the redundant constraints, we make \(\mathcal{F}_\alpha\) the first frame observing \(S_j\).
This reprojection among keyframes induced by a surfel is shown in Fig. \ref{fig:surfel_reprojection}.
Since in Equation \ref{eqn:surfel_reproj_res}, \(\mathbf{x}_W\) can be presented by \(\mathbf{T}^W_k\), \(\mathbf{p}_k\) and \(\bm{\omega}_S\), the variables to be optimized in the final cost function, Equation \ref{eqn:surfel_reproj_cost}, remain the keyframe poses \(\mathbf{T}^W_{(\cdot)}\) only, which can be efficiently solved.
After obtaining the optimized keyframe poses, we can further update the map point position by Equation \ref{eqn:world_point} with the first observed keyframe pose.

\subsection{3D Geometry-Aided Visual Relocalization} \label{sec:hierarchical_relocal}
As shown in Fig. \ref{fig:relocalization_diagram}, we adopt a hierarchical relocalization scheme, containing database frame detection and 6-DoF pose estimation, to perform the relocalization efficiently.
Firstly, we retrieve database keyframes which have the most overlapping visual content with the current frame.
The global feature extractor will extract a global descriptor from the new query image.
Keyframes from the visual database are detected based on the similarity of the global descriptor of the query image and those of database frames.
Our database returns the top-\(K\) matched keyframes, which are clustered by the covisibility information, i.e. whether the matched keyframes share the same map points.
This procedure returns \(N_{db}\) clusters.
We then take the keyframe with the highest similarity score in each cluster as the canonical keyframe \(\mathcal{F}_{c}\).
To reduce the final candidate database keyframes and limit the relocalization time, we keep \(\min(N_{db}, N_{\max})\) canonical keyframes to process the following camera pose estimation.

For each canonical keyframe \(\mathcal{F}_{c}\), we find \(N_{co}\) covisible frames \(\{\mathcal{F}\}_{co}\) with highest number of shared map points.
The map points visible in \(\mathcal{F}_{c} \cup \{\mathcal{F}\}_{co}\) are matched with 2D feature points in the query frame, providing 2D-3D matches.

Because pose errors exist in the database building step, some map points may not be considered as visible or matchable in \(\mathcal{F}_{c} \cup \{\mathcal{F}\}_{co}\).
Only using the map points visible in the local keyframes \(\{ \mathbf{x}\}^F_c\) might lead to too few good matches, which may cause the PnP solver to be unreliable.
Thus, we consider using projected neighboring map points to increase the number of matches.
We collect the map points not in \(\{ \mathbf{x}\}^F_c\) but associated with \(\{S\}^n_c\), the neighboring surfels of \(\mathcal{F}_{c}\), as \(\{ \mathbf{x}\}^N_c\).
Another matching between \(\{ \mathbf{x}\}^N_c\) and the non-matched 2D feature points is then performed to find more correspondences.

To reduce mismatches, an essential matrix check between \(\mathcal{F}_c\) and \(\mathcal{F}_{q}\) is performed before passing the matches to the pose solver.
If the matched map points are visible in \(\mathcal{F}_c\), the corresponding keypoint positions are used in \(\mathcal{F}_c\); if they are not considered as visible, we project the map point positions to \(\mathcal{F}_c\) to obtain the keypoint positions.
The projected keypoints may be out of \(\mathcal{F}_c\).
Instead of throwing away the outside points, we use them in the essential matrix check and keep them if they are considered as inliers after the check.
This helps us to retain the map points that are not visible in \(\mathcal{F}_c\) but matchable in \(\mathcal{F}_q\).

With the verified 2D-3D matches, we perform PnP with RANSAC \cite{lepetit2009epnp} to find a rough camera pose for each \(\mathcal{F}_c\) containing enough matches.
After obtaining the rough pose, we project the local 3D map points to the current image plane again, and find more correspondences in the local grid of the query image.
These 2D-3D matches construct a motion-only optimization problem \cite{mur2017orb} based on reprojection errors, as shown in Equation \ref{eqn:motion_ba}:
\begin{equation}\label{eqn:motion_ba}
    \begin{aligned}
        \mathbf{e}_i &= \mathbf{p}_i - \Pi(\mathbf{T}^W_q, \mathbf{x}_{W,i}) \\
        c_r &= \sum_i{\rho(\mathbf{e}_i^\top \mathbf{\Omega}^{-1}_i \mathbf{e}_i)}
    \end{aligned},
\end{equation}
where \(\mathbf{T}^W_q\) is the pose of the query image, \(\Pi\) is the camera projection function, \(\mathbf{p}_i\) is the \(i^\text{th}\) keypoint position, \(\mathbf{x}_{W,i}\) is the matched map point, \(\rho\) is the Huber loss function, and \(\mathbf{\Omega}_i\) is the covariance matrix from the keypoint scale.
Minimizing \(c_r\) is a non-linear optimization problem, which can be efficiently solved by Ceres Solver \cite{ceres-solver}.

A pose validation step is used to filter out the wrong poses.
We first check the inliers of keypoint-to-map-point (2D-3D) matches provided by the optimized current pose, \(n_{in}\).
If \(n_{in}\)\(\ge\)\(\theta_{in}\), where \(\theta_{in}\) is a threshold for inliers, the pose is regarded as an inlier pose.
Then we check the distance between the current inlier pose and the last inlier pose, \(d_{in}\).
If \(d_{in}\)\(\le\)\(\theta_{dist}\), the current inlier pose is verified and the relocalization system returns a valid final pose for the current query image.

\section{Results} \label{sec:s4_results}

In this section, we evaluate the proposed method with real-world and simulation tests.
An ablation study is also conducted to show the effectiveness of our design.

\subsection{Implementation Details}
We compare two local feature extractors, ORB \cite{rublee2011orb} and SuperPoint (SP) \cite{detone2018superpoint}.
For fair comparison with our ORB baseline method, ORB-SLAM \cite{mur2015orb}, we use the local feature detection module from ORB-SLAM, which extracts homogeneously distributed FAST corners \cite{rosten2006machine} among different pyramids of the input images, and the same ORB descriptors \cite{rublee2011orb} for the detected keypoints.
A total 1000 ORB keypoints are extracted in each image.

For global feature extractors, we compare two representations of images, the reduced VLAD vector in NetVLAD \cite{arandjelovic2016netvlad} and bag of binary words (BoBW) \cite{galvez2012bags}.
A KD-tree structure of global NetVLAD descriptors is formed for efficient similarity score calculation and the inverted index \cite{galvez2012bags} is used in the BoBW database.

We set the top-\(K\)=30, \(N_{\max}\)=3 and \(N_{co}\)=5 to balance the relocalization speed and accuracy.
For the pose verification, \(\theta_{in}\)=15, and \(\theta_{dist}\)=0.3 m for indoors and \(\theta_{dist}\)=0.8m for outdoors are set in all tests, unless specified otherwise.

\subsection{EuRoC Dataset}
For the real-world indoor experiments, we use the EuRoC dataset \cite{burri2016euroc}, which provides grayscale stereo images captured on a UAV, IMU data, ground-truth poses and a ground-truth lidar point cloud map.
In our tests, the ground-truth lidar point cloud map is used to build a surfel map with surfel radius 0.02 m, as described in \cite{ye2020monocular}.
For visual database building, the camera poses in the surfel map are estimated by our previous method DSL \cite{ye2020monocular}.

We take the monocular images from sequences V102 and V202 to build the visual database of two different rooms V1 and V2, respectively.
The sequences V101, V103 are used to test the relocalization performance for room V1; and V201 and V203 are used for the testing of room V2.

We compare two variants with SP or ORB as local feature extractors, ORB-SLAM \cite{mur2015orb} and NV+SP \cite{sarlin2019coarse}.
NV+SP is the version in \cite{sarlin2019coarse} taking NetVLAD for global retrieval and SuperPoint for local feature extraction.
It requires an offline SfM process to build a 3D model to provide covisibility information for exhaustive inter-frame feature matching.
Both ORB-SLAM \footnote{To perform the relocalization with a single image input, ORB-SLAM is set to the relocalization mode without the function of visual odometry.} and our method take sequential image inputs to build the visual database.
They both compute the covisibility information on the fly and are more flexible for updating the visual database.
ORB-SLAM estimates keyframe poses and map points up to scale.
Hence we align the database keyframe trajectory of ORB-SLAM to the ground camera trajectory according to \cite{umeyama1991least}.
Our method uses 3D information from a surfel map, which makes the keyframe poses and map points consistent with the metric environment.

The pose recall at position threshold \(\theta_r = 0.3 \ \text{m}\) and average absolute trajectories error (mATE) \cite{sturm2012benchmark} in centimeters of the relocalized poses for the different methods are shown in Table \ref{tab:euroc_relocalization}.
Note that NV+SP \cite{sarlin2019coarse}, which uses accurate SfM models \cite{schoenberger2016sfm} to perform localization, has the highest recall in all tests since the covisibility information and feature matching are more exhaustively examined than those from the methods taking sequential images as input. 
However, it is time-consuming to build and update such SfM models.
Our method outperforms ORB-SLAM in terms of recall and has comparative recall with NV+SP in easy sequences V101 and V201.
Since the ORB extractor extracts more feature points (\(\sim 1000\)) than the SuperPoint extractor (\(\sim 500\)), it gives the system slightly higher recall in easy sequences.
However, in difficult sequences, where lighting conditions and view-points are changed greatly, the SuperPoint extractor can provide more consistent feature detection and description than ORB extractor.
Thus, the SP variant of our method has a much higher recall.
It is worth noting that with pose verification, our method has the lowest mATE in most of the tests.
We notice that NV+SP has large pose outliers in difficult sequences V103 and V203.
Thus we only show the mATE of NV+SP's poses whose errors are less than \(5 \ \text{m}\).

\begin{figure}[ht]
  \centering
  \includegraphics[width=0.35\textwidth]{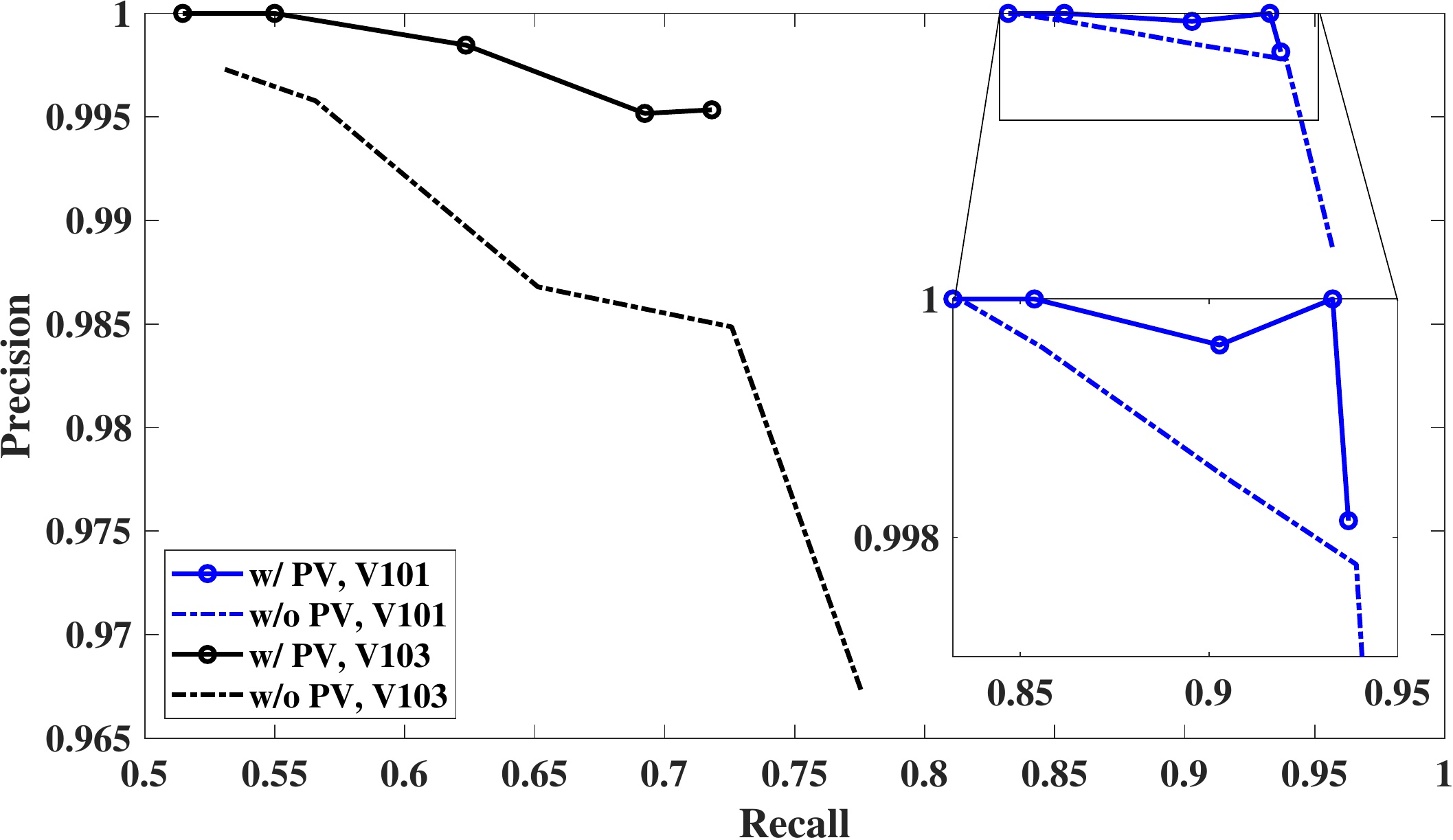}
  \caption{Precision-recall curves achieved by our method with/without pose verification (PV).
  }
  \label{fig:euroc_pr}
\end{figure}

To further show the effectiveness of the pose verification introduced in Sec. \ref{sec:hierarchical_relocal}, we draw the precision-recall curves by changing the numbers of inliers to determine the success of relocalization in Fig. \ref{fig:euroc_pr}.
With pose verification, most false relocalization results can be detected and filtered.
This leads to higher the precision of the final relocalization results, even in the difficult V103 sequence.

\begin{table}[!t]
  \centering
  \caption{Evaluation of the relocalization on the EuRoC dataset by different methods.
  Recall and average ATE (mATE) with unit in [cm] are shown with the format (\% / cm).
  The best and the second-best results are in bold and with underlines, respectively.
  V102 and V202 are used for building visual databases for the V1 and V2 tests.
  The proposed method and ORB-SLAM \cite{mur2015orb} use images sequentially to build the visual database, while NV+SP \cite{sarlin2019coarse} matches all images to build the visual database by SfM \cite{schoenberger2016sfm}.
  }
  \begin{tabular}{lcccc}
    \toprule
          & Ours (SP) & Ours (ORB) & ORB-SLAM & NV+SP \\
    \midrule
    V101  & 93.5 / 4.91 & \underline{94.2} / 4.49 & 55.8 / \underline{3.11} & \textbf{99.9} / \textbf{2.50} \\
    V103  & \underline{71.5} / \textbf{5.25} & 61.4 / \underline{5.95} & 17.6 / 13.28 & \textbf{93.8} / 20.44 \\
    V201  & 88.3 / \textbf{2.82} & \underline{90.1} / 3.41 & 61.5 / \underline{2.96} & \textbf{98.9} / 5.54 \\
    V203  & \underline{53.5} / \textbf{5.20} & 26.8 / 10.29 & 1.0 / \underline{9.85} & \textbf{88.7} / 32.39 \\
    \bottomrule
    \end{tabular}%
  \label{tab:euroc_relocalization}%
\end{table}%

\begin{table}[htbp]
  \caption{Ablation study on the EuRoC dataset, with VX02 used for building the visual database, and VX01 and VX03 for testing.
  Relocalization recall and average ATE in [cm], (\% / cm), with different local descriptors, SuperPoint (top), ORB (bottom), and other designs of the proposed relocalization approach are shown.}
  \vspace{-0.5em}
  \begin{tabular}{cc}
    \begin{tabular}{lccc}
      \hline
      \multicolumn{4}{c}{SP} \bigstrut\\
      \hline
      Global & NV    & NV    & NV \bigstrut[t]\\
      Local & FN    & F     & Naive \bigstrut[b]\\
      \hline
      V101  & \textbf{93.5} / \textbf{4.91} & 90.4 / 5.00 & 86.1 / 6.50 \bigstrut[t]\\
      V103  & \textbf{71.5} / \textbf{5.25} & 65.2 / 5.38 & 54.0 / 9.87 \\
      V201  & \textbf{88.3} / \textbf{2.82} & 84.7 / 2.90 & 83.8 / 4.30 \\
      V203  & \textbf{53.5} / \textbf{5.20} & 47.6 / 5.03 & 45.1 / 6.57 \bigstrut[b]\\
      \hline
      \end{tabular}%
    \\
    \vspace{-0.5em}
    \\
    \begin{tabular}{lcccc}
      \hline
      \multicolumn{5}{c}{ORB} \bigstrut\\
      \hline
      Global & NV    & NV    & NV    & BoBW \bigstrut[t]\\
      Local & FN    & F     & Naive & FN \bigstrut[b]\\
      \hline
      V101  & \textbf{94.2} / \textbf{4.49} & 87.0 / 4.60 & 80.3 / 5.84 & 88.7 / 4.64 \bigstrut[t]\\
      V103  & \textbf{64.1} / 5.95 & 56.3 / \textbf{5.32} & 42.6 / 6.85 & 59.8 / 6.95 \\
      V201  & \textbf{90.1} / 3.41 & 85.6 / 3.64 & 80.4 / 4.15 & 89.4 / \textbf{3.31} \\
      V203  & \textbf{26.8} / 10.29 & 19.8 / 10.97 & 11.9 / 15.20 & 18.0 / \textbf{8.98} \bigstrut[b]\\
      \hline
      \end{tabular}%
  \end{tabular}
  \label{tab:relocalization_ablation}%
\end{table}%

Table \ref{tab:relocalization_ablation} shows an ablation study of the proposed method using different designs of the system.
Relocalization recall at \(\theta_r = 0.3\text{m}\) and mATE in [cm] are shown.
For global retrieval, NV denotes NetVLAD \cite{arandjelovic2016netvlad} and BoBW denotes the BoBW module similar to that in \cite{galvez2012bags,mur2015orb} for ORB features.
Different points used in our local matching and pose estimation are denoted as local visible points \(\{\mathbf{x}\}^F_c\) (F) and neighboring surfel-related map points \(\{\mathbf{x}\}^N_c\) (N), as introduced in Sec. \ref{sec:hierarchical_relocal}.
The direct association of the keypoints and the projected 3D surfel centers is denoted as (Naive) with the best matched keyframe.
It does not consider the covisibility information nor map point fusion and culling in the visual database.
The results show that the global retrieval based on NetVLAD descriptors obtains higher recall than the BoBW-based variant.
Since our method involves covisibility information and keeps map points trackable among multiple frames, when performing the 6-DoF camera pose estimation, it has a higher recall and a lower mATE compared to the Naive method.
Neighboring surfel-related map points \(\{\mathbf{x}\}^N_c\) further improve the relocalization performance by providing more 2D-3D matches.

\subsection{CARLA Simulator}
In this section, we evaluate our method in the CARLA simulator \cite{dosovitskiy2017carla}, which generates image inputs, lidar inputs and ground truth poses.
Similar to \cite{ye2020monocular}, a 3D surfel map is built by point cloud map from lidar inputs and poses with a 0.1 m average surfel radius.

Visual databases are built by different sources of camera poses, as discussed in Sec. \ref{sec:database}:
camera poses from a localization algorithm (DSL) \cite{ye2020monocular} and those transformed from lidar poses with extrinsic parameters between the lidar and camera.
We show how the perturbations of camera poses affect the relocalization performance.

In the visual database building step, we add random noises with standard deviation \(\sigma\) to the positions of the camera pose to simulate the perturbations of extrinsic parameters and time offsets between the lidar and camera.
After building the visual database, we use the sequence with a car driving in different lanes to test the relocalization.
The relocalization recall and average ATE are shown in Table \ref{tab:carla_pose_noise}.
From the results, we can see that, with larger pose errors, the performance of relocalization becomes worse.
Our optimization introduced in Sec. \ref{sec:surfel_reproj_opt} helps to improve the quality of the visual database.
With optimization, the recall increases, and mATE drops, especially for the noisier camera poses.

\begin{table}[htbp]
  \vspace{-0.5em}
  \centering
  \caption{Relocalization results in CARLA simulator dataset with poses transformed from noisy lidar poses or DSL \cite{ye2020monocular}.
  The statistics of the corresponding visual databases, relocalization recall and mATE \cite{sturm2012benchmark} of relocalization are shown.
  Optimization denotes the visual database optimization introduced in Sec. \ref{sec:surfel_reproj_opt}.
  }
  \begin{tabular}{l|c|C{0.4cm}C{0.4cm}|C{0.4cm}C{0.4cm}||C{0.4cm}C{0.4cm}}
    \hline
    \multirow{2}[4]{*}{\shortstack[l]{Source of \\ camera poses}} & \multicolumn{5}{c||}{Transformed camera poses} & \multicolumn{2}{c}{\multirow{2}[4]{*}{DSL poses}} \bigstrut\\
    \cline{2-6}      & \(\sigma=0.0 \text{m}\) & \multicolumn{2}{c|}{\(\sigma=0.2 \text{m}\)} & \multicolumn{2}{c||}{\(\sigma=0.4 \text{m}\)} & \multicolumn{2}{c}{} \bigstrut\\
    \hline
    \# of map point & 25683 & \multicolumn{2}{c|}{22543} & \multicolumn{2}{c||}{18315} & \multicolumn{2}{c}{31080} \bigstrut[t]\\
    \# of keyframes & 323   & \multicolumn{2}{c|}{501} & \multicolumn{2}{c||}{559} & \multicolumn{2}{c}{564} \bigstrut[b]\\
    \hline
    Optimization & N/A   & w/o   & w/    & w/o   & w/    & w/o   & w/ \bigstrut\\
    \hline
    Recall [\%] & 81.1  & 78.8  & 82.0  & 39.3  & 74.9  & 83.7  & 84.7 \bigstrut[t]\\
    mATE [cm] & 16.7  & 49.2  & 39.2  & 77.0  & 52.7  & 18.4  & 17.8 \bigstrut[b]\\
    \hline
    \end{tabular}%
  \label{tab:carla_pose_noise}%
  \vspace{-1em}
\end{table}%

\begin{figure}[ht]
  \centering
  \begin{subfigure}[c]{0.21\textwidth}
		\centering
        \includegraphics[width=\textwidth]{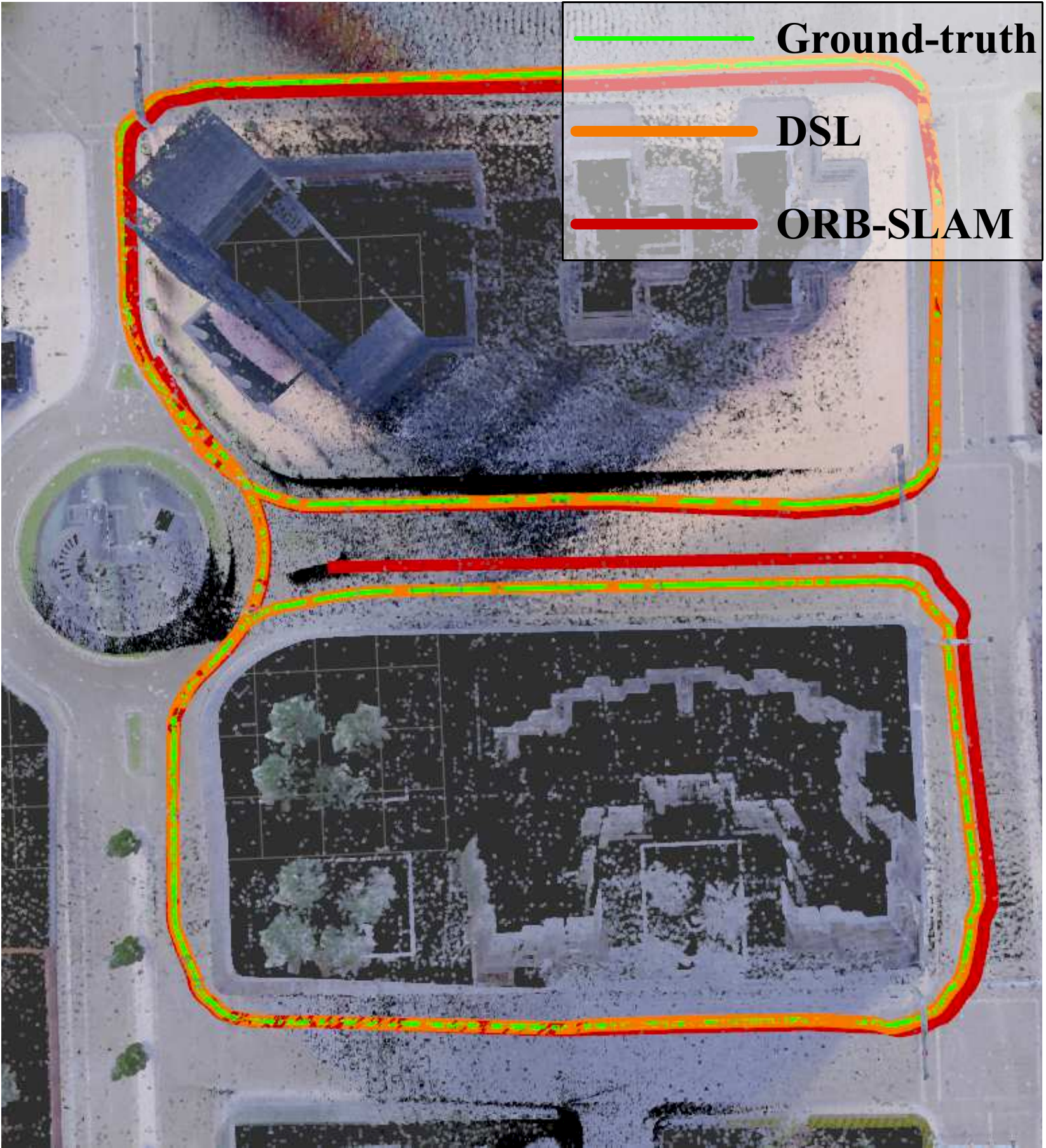}
        \caption{Database trajectories}
		\label{fig:database_carla}
	\end{subfigure}
	\begin{subfigure}[c]{0.2\textwidth}
		\centering
        \includegraphics[width=\textwidth]{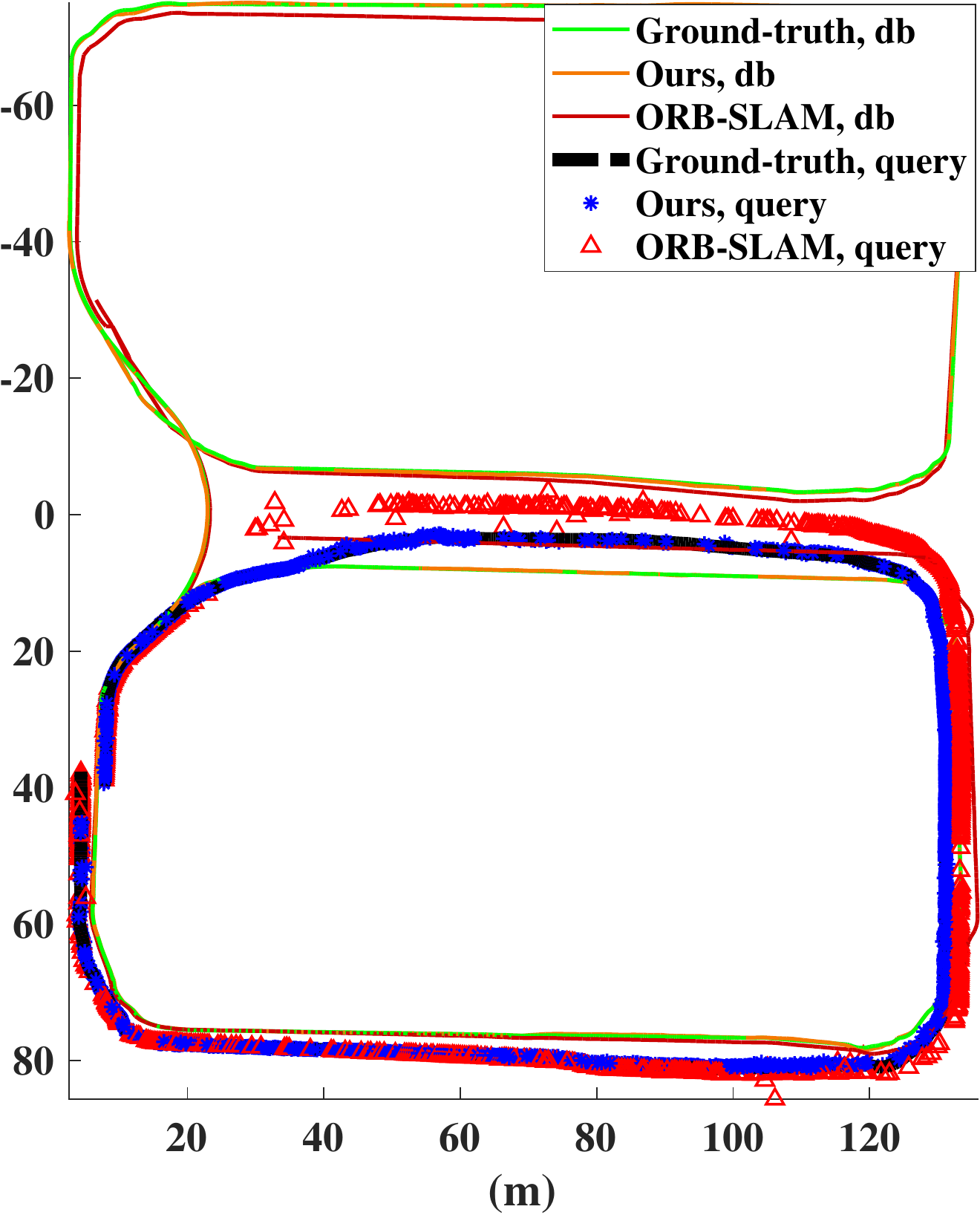}
        \caption{Relocalization results.}
		\label{fig:query_carla}
	\end{subfigure}
  \caption{Map building and relocalization results on CARLA simulation.
  (a) Different trajectories for database building.
  (b) The trajectory for database building is denoted as ``db''; the relocalization results of one query sequence is denoted as ``query''.
  Since our method takes 3D geometric information in the ``db'' step, metrically correct relocalization results can be provided in the ``query'' time.}
  \label{fig:carla_relocalization}
  \vspace{-1em}
\end{figure}
Fig. \ref{fig:carla_relocalization} shows the relocalization results of our method and ORB-SLAM.
In our test, ORB-SLAM had scale-drift when building the visual map.
Even if we align the camera trajectory \cite{umeyama1991least} to the ground-truth, the drift from the visual system cannot be eliminated completely.
This causes large relocalization errors in the regions where the visual system drifts, since the database itself is not well aligned with the 3D environment.
Our method takes 3D surfel information in the map building step, and thus can provide metrically correct relocalization results.

\subsection{Runtime}
Runtime analysis
\footnote{Run
on an Intel i7-8700K CPU with an Nvidia GTX-1080Ti GPU.}
for the different steps of our method are shown in Table \ref{tab:relocalization_time}.
The mean process time of the visual database building step is around 50--100 ms, and the relocalization requires less than 50 ms if we extract the local and global descriptors in another independent thread.
The visual database optimization runs only once after the sequential images are processed by the visual database updater.
Thus, we show the time divided by the number of input images.

We also follow the same procedures as described in \cite{sarlin2019coarse} to build the visual database for NV+SP.
The database building takes more than 20 times the duration of the input sequence, which includes taking 600--900 keyframes for COLMAP \cite{schoenberger2016sfm} to build the SfM model and using the SfM results to find SuperPoint matches and perform triangulation to build a visual database using SuperPoint.
Avoiding building an SfM model, our method is more efficient at the time of building the visual database and the visual database can be updated by new sequential images and camera trajectories.

\begin{table}[htbp]
\centering
\vspace{-0.5em}
\caption{Runtime of visual database building and relocalization per input image.}
\begin{tabular}{ll|cc||cc}
  \hline
  \multicolumn{2}{l|}{\textbf{Runtime [ms]}} & \multicolumn{2}{c||}{\textbf{EuRoC}} & \multicolumn{2}{c}{\textbf{CARLA}} \bigstrut\\
  \hline
  \multicolumn{2}{l|}{Local descriptor} & SP    & ORB   & SP    & ORB \bigstrut\\
  \hline
  \multicolumn{2}{l|}{Feature extraction} & 16.1  & 12.7  & 20.9  & 14.2 \bigstrut\\
  \hline
  \multicolumn{2}{p{10.36em}|}{Visual database building} & 47.7  & 103.4 & 85.7  & 116.6 \bigstrut[t]\\
        & Map rendering \& association & 14.6  & 19.8  & 20.3  & 25.0 \\
        & Visual database updater  & 31.2  & 75.3  & 60.6  & 88.3 \\
        & Visual database optimization & 1.9   & 8.3   & 4.8   & 3.3 \bigstrut[b]\\
  \hline
  \multicolumn{2}{l|}{Global descriptor (NV)} & \multicolumn{2}{c||}{25.6} & \multicolumn{2}{c}{31.9} \bigstrut\\
  \hline
  \multicolumn{2}{l|}{Relocalization} & 30.6  & 47.8  & 44.3  & 47.7 \bigstrut[t]\\
        & Global retrieval & \multicolumn{2}{c||}{2.1} & \multicolumn{2}{c}{2.8} \\
        & Local pose estimation & 28.5  & 45.7  & 41.6  & 44.9 \bigstrut[b]\\
  \hline
  \multicolumn{2}{l|}{Image size} & \multicolumn{2}{c||}{752 \(\times\) 480} & \multicolumn{2}{c}{800 \(\times\) 600} \bigstrut\\
  \hline
  \end{tabular}%
  \label{tab:relocalization_time}%
  \vspace{-2em}
\end{table}%

\section{Conclusion} \label{sec:conclusion}
In this work, we have introduced a camera relocalization algorithm which utilizes a 3D surfel map to build a visual database for relocalization.
The proposed method utilizes the surfel map rendering to associate map points with surfels to create covisible information.
Surfel constraints among frames are further utilized to improve the quality of the visual database.
Given a query image, a hierarchical relocalization is performed with detection of the closest database keyframes and the 6-DoF pose estimation.
Both real-world and simulation tests show that our method takes 3D geometric information from the pre-built surfel map to achieve recall comparable with SfM model-based approaches and provide metric relocalization results consistent with the 3D environment.
Our future work includes investigating the online updating of the visual database and 3D structure to improve the relocalization ability and handling relocalization in more dynamic environments.

\addtolength{\textheight}{-3.65cm}   %

\bibliographystyle{IEEEtran}
\bibliography{root_bib}

\end{document}